\documentclass[sigconf]{acmart}
\usepackage{booktabs} 
\usepackage{enumitem}
\usepackage{amsmath}
\usepackage{amsfonts}
\usepackage[ruled,vlined]{algorithm2e}
\usepackage{algpseudocode}
\usepackage{bm}
\usepackage{subfigure}
\usepackage{multirow}
\usepackage{hyperref}
\usepackage{fancyhdr}

\algnewcommand{\LeftComment}[1]{ \(\triangleright\) #1}

\setcopyright{none}
\copyrightyear{}
\acmYear{}
\acmDOI{}
\renewcommand\footnotetextcopyrightpermission[1]{}

\acmConference[TheWebConf'23]{The 2023 ACM Web Conference}{April 30 -- May 4, 2023}{Austin, Texas}
\acmPrice{15.00}
\acmISBN{978-1-4503-XXXX-X/18/06}

\begin{document}

\title{Toward Robust Graph Semi-Supervised Learning against Extreme Data Scarcity}


\author{Kaize Ding$^\dagger$,~~~Elnaz Nouri$^\ddagger$,~~~Guoqing Zheng$^\ddagger$,~~~Huan Liu$^\dagger$, ~~~Ryen White$^\ddagger$}

\affiliation{
 \institution{$^\dagger$Arizona State University, \{kaize.ding, huan.liu\}@asu.edu \\ $^\ddagger$Microsoft Research, \{Elnaz.Nouri, Guoqing.Zheng, Ryen.White\}@microsoft.com}
  \country{}
}

\begin{abstract}
    
The success of graph neural networks on graph-based web mining highly relies on abundant human-annotated data, which is laborious to obtain in practice. When only few labeled nodes are available, how to improve their robustness is a key to achieve replicable and sustainable graph semi-supervised learning. Though self-training has been shown to be powerful for semi-supervised learning, its application on graph-structured data may fail because (1) larger receptive fields are not leveraged to capture long-range node interactions, which exacerbates the difficulty of propagating feature-label patterns from labeled nodes to unlabeled nodes; and (2) limited labeled data makes it challenging to learn well-separated decision boundaries for different node classes without explicitly capturing the underlying semantic structure. To address the challenges of capturing informative structural and semantic knowledge, we propose a new graph data augmentation framework, AGST (\underline{A}ugmented \underline{G}raph \underline{S}elf-\underline{T}raining), which is built with two new (i.e., structural and semantic) augmentation modules on top of a decoupled GST backbone. In this work, we investigate whether this novel framework can learn a robust graph predictive model under the low-data context. We conduct comprehensive evaluations on semi-supervised node classification under different scenarios of limited labeled-node data. The experimental results demonstrate the unique contributions of the novel data augmentation framework for node classification with few labeled data. 



    
    
\end{abstract}


\maketitle
\pagestyle{plain}



\section{Introduction}


With the rapid development of the World Wide Web, recent years we have witnessed the growth in our ability to generate and gather data on numerous online and offline platforms. Graphs, where entities are denoted as nodes and the relations connecting them are denoted as edges, have become a common language for modeling a plethora of structured and relational systems on the web, ranging from social networks~\cite{zafarani2014social}, to knowledge graphs~\cite{wang2017knowledge}, to e-commerce user-item interaction graphs~\cite{mcauley2015inferring}. To ingest the valuable information encoded in graph-structured data, graph learning algorithms have been proposed in the research community and made huge success in different domains. Recently, graph neural networks (GNNs), a generalized form of neural networks for graph-structured data, have become the prevailing paradigm due to their effectiveness and scalability~\cite{kipf2017semi,velickovic2018graph,hamilton2017inductive}.



As a typical graph-related web mining task, node classification has received continuous endeavors in the World Wide Web community. Existing GNNs developed for node classification usually focus on the canonical \textit{semi-supervised} setting where relatively abundant gold-labeled nodes are provided. While this setting is often impractical since data labeling is extremely labor intensive, especially when considering the heterogeneity of graph-structured data~\cite{yao2020graph,ding2020graph}. To overcome the data scarcity issue, self-training or pseudo-labeling~\cite{lee2013pseudo} has been explored to combine with GNNs and proven to be effective for solving semi-supervised node classification with fewer labels~\cite{li2018deeper,sun2020multi,ding2022data}.




Existing GST methods, however, simply combine the idea of self-training with GNNs, which can be ineffective in handling graph data with few labeled nodes, or in exploiting numerous unlabeled nodes due to two limitations: (1) \textbf{\textit{Structural Bottleneck.}} Given few labeled nodes, it is important for the GNN model to enable more propagation steps so the feature patterns of labeled nodes can be better propagated to the long-distance unlabeled nodes. However, recent work pointed out the distortion of information flowing from distant nodes (i.e., over-squashing~\cite{alon2021bottleneck,topping2022understanding}) as a factor limiting the efficiency of message-passing for tasks relying on long-range node interactions. In addition, real-world graphs often come with a certain level of structure noise (e.g., ``noisy'' and ``missing'' edges), which could be generated by either adversaries~\cite{zhu2021deep} or the data collection process itself~\cite{zhao2021data}. Such structure noise can easily interfere with the message-passing process and make it difficult to learn correct feature-label patterns with few labeled nodes. Hence, it is crucial to avoid over-squashing and reduce data noise in our endeavor to further improve the performance of GST 
with few labeled nodes; and (2) \textbf{\textit{Semantic Bottleneck.}} It is challenging to learn well-separated decision boundaries between different node classes when labeled training data is severely scarce and the semantic manifold is complex. Though GST methods attempt to alleviate the data scarcity problem by adding pseudo labels, the nodes with pseudo-labels may introduce complex feature patterns and pseudo labels can be unreliable, which causes the model performance to deteriorate. We ask if novel ideas can be explored to optimize the usage of pseudo labels for capturing the underlying semantic structure of the sparsely-labeled graph.

In this paper, we propose an \underline{A}ugmented \underline{G}raph \underline{S}elf-\underline{T}raining framework, namely AGST, for tackling semi-supervised node classification 
where few labeled nodes are available. 
We plan to address the limitations of conventional GST methods by proposing two original modules for  \textit{structural and semantic data augmentations}. Specifically, our framework employs a simple, decoupled GNN as the GST backbone, where the teacher model first performs high-order label propagation to generate pseudo labels on unlabeled nodes based on Personalized PageRank~\cite{page1999pagerank}, and the student model conducts feature transformation by mapping the features of nodes to their gold/pseudo labels. From the \textbf{\textit{structural data augmentation}} perspective, our framework not only enables large receptive fields to capture long-range node interactions, but also avoids the over-squashing issue by decoupling the transformation and propagation steps in message passing. To further promote the information propagation, in each GST iteration, a deterministic topology augmentation function is derived from the learned model and is utilized to refine the input topological structure for the next iteration. This way we expect that AGST can capture richer (i.e., both local and global) and cleaner (i.e., less noisy) structure knowledge during the GST process. From the \textbf{\textit{semantic data augmentation}} perspective, the pseudo labels introduced by the teacher model can enrich the semantic knowledge of the training data when learning the student model. To optimize the semantic alignment between the few labeled nodes and generated pseudo-labeled nodes, we suggest to explicitly capture the semantic structures of the input graph by proposing a weakly-supervised contrastive loss to encourage intra-class compactness and inter-class separability in the latent feature space. As such, well-separated decision boundaries can be learned during the GST process even with few labeled nodes. The proposed AGST framework enables the two data augmentation modules to work seamlessly with the decoupled GST backbone and to quickly learn a robust graph predictive model even with few labeled nodes. To summarize, our key contributions are listed as follows:

\begin{itemize}[leftmargin=*,noitemsep,topsep=2.5pt]

\item \textbf{\emph{Problem}}: We investigate the problem of semi-supervised node classification under the challenging low-data setting, which focuses on improving the replicability and sustainability of GNNs in practical scenarios.
    
\item \textbf{\emph{Algorithm}}: We propose a principled GST framework, which differs from the existing efforts and improves the performance with scarce labeled data by augmenting data from both structural and semantic perspectives.
    
\item \textbf{\emph{Evaluation}}: We conduct extensive experiments on various real-world datasets to evaluate the effectiveness of our approach. The experimental results demonstrate the unique contributions made by AGST to performance improvement over existing methods.

\end{itemize}

\section{Related Work}
 \begin{figure*}[!t]
    \graphicspath{{figures/}}
    \centering
    \includegraphics[width=0.95\textwidth]{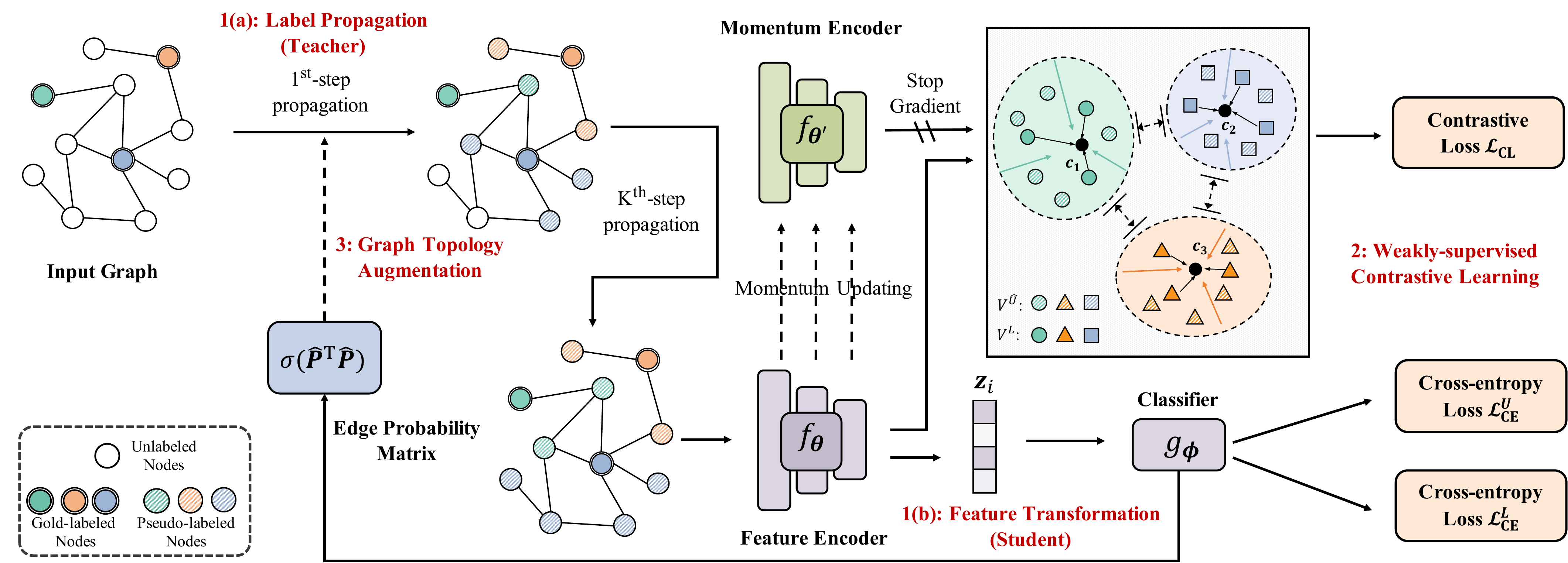}
    \caption{Overview of the proposed framework. In each training iteration, AGST first perform high-order label propagation to generate pseudo labels on unlabeled nodes, then conduct feature transformation with the augmented training label set. In the meantime, a weakly-supervised contrastive loss is used for optmizing the usage of pseudo labels. Based on the computed edge probability matrix, the input graph structure will be augmented and fed to the next iteration. Figure best viewed in color.}%
    \label{fig:framework}%
\end{figure*}

\smallskip
\noindent\textbf{Graph Neural Networks.} Graph neural networks (GNNs), a family of neural endeavors for learning latent node representations on a graph, have drawn much attention in the community of graph machine learning (GML)~\cite{bruna2013spectral,defferrard2016convolutional,kipf2017semi,velickovic2018graph}. In general, graph neural networks can be categorized into spectral~\cite{bruna2013spectral,defferrard2016convolutional,kipf2017semi,wu2019simplifying} and spatial approaches~\cite{hamilton2017inductive,velickovic2018graph,xu2018powerful}. Originally inspired by graph spectral theory, spectral-based graph convolutional networks (GCNs) extend convolution operation in the spectral domain to graph-structured data. Among them, the model proposed by Kipf et al.~\cite{kipf2017semi} has become the most prevailing one by using a linear filter. Later on, SGC~\cite{wu2019simplifying} is proposed to further reduce the computational complexity by removing the non-linearity of GCNs. As another line of work, spatial-based GNNs define graph convolutions based on a node’s spatial relations~\cite{hamilton2017inductive,velickovic2018graph,xu2018powerful}. For example, GAT~\cite{velickovic2018graph} incorporates trainable attention weights to specify fine-grained weights on neighbors when aggregating neighborhood information of a node. In essence, although spectral-based and spatial-based GNNs start on a different basis, both of them share the same propagation rule, which is the message-passing scheme. Those methods model the homophily principle~\cite{mcpherson2001birds} and learn node representations by iteratively transforming, and propagating/aggregating node features within graph neighborhoods. When long-range node interactions are needed, the over-squashing issue can largely undermine the model performance if we directly increase the model depth. Thus researchers try to solve this issue by proposing different techniques, such as sampling or rewiring edges~\cite{hamilton2017inductive,monti2018motifnet}, decoupling the feature transformation and propagation steps~\cite{klicpera2019predict,huang2020graph,liu2020towards,dong2021equivalence,chien2021adaptive} and many others~\cite{li2018deeper,xu2018representation}. In particular, decoupled GNNs~\cite{dong2021equivalence} have become a prevailing paradigm due to their simplicity and learning efficiency. For example, APPNP~\cite{klicpera2019predict} propagates the neural predictions via personalized PageRank, which can preserve the node’s local information while increasing the receptive fields. Liu et al.~\cite{liu2020towards} propose to decouple the propagation and transformation steps and then utilizes an adaptive adjustment mechanism to balance the information from local and global neighborhoods of each node. However, the aforementioned models neglect the additional supervision signals from unlabeled data, which has become a bottleneck for pushing the performance boundary of GNNs. Though CGPN~\cite{wan2021contrastive} leverages poisson learning to propagate the labels to the entire graph, it cannot address the structure noise and explicitly capture the semantic structures of the input graph.




\smallskip
\noindent\textbf{Node Classification with Few Labels.}
In real-world scenarios, labeled training samples are usually quite limited due to the intensive cost of data labeling. Albeit the great success of GNNs for graph-based semi-supervised learning, most of the existing efforts are designed as shallow models with a restricted receptive field, leading to their ineffectiveness in limited labeled data scenarios~\cite{liu2020towards,dong2021equivalence,lin2020shoestring}. Under the extreme cases when very few labels are given, shallow GNNs cannot effectively propagate the training labels and characterize the global information of the input graph~\cite{li2018deeper}. Many advanced deep GNNs~\cite{klicpera2019predict,liu2020towards,xu2018representation,chien2021adaptive,yang2021extract} have shown their advantages in leveraging large receptive fields for propagating label signals. However, the main concern in semi-supervised node classification, i.e., the shortage of supervision information, has not been directly addressed. To counter this issue, self-training~\cite{yarowsky1995unsupervised}, also known as pseudo-labeling~\cite{lee2013pseudo}, where one imputes labels on unlabeled data based on a teacher model trained with limited labeled data, has been applied to improve GNNs to solve the problem of semi-supervised node classification. Among those methods, Li et al.~\cite{li2018deeper} first combine GCNs and self-training to expand supervision signals. Furthermore, M3S~\cite{sun2020multi} proposed multi-stage self-training and utilized clustering method to eliminate the pseudo labels that may be incorrect. Similar ideas can also be found in \cite{zhou2019effective,dai2021nrgnn,liu2022confidence}. However, existing methods still adopt shallow GNNs to build the teacher and student models, which inherently restrict the effective propagation of label signals. Apart from the aforementioned methods, our AGST framework adopts a decoupled design, in which the teacher model is a label propagation module and the student model is a feature transformation module. As such, our framework is capable of leveraging both large receptive fields and pseudo supervision signals, making the learned model more label-efficient. In addition, we propose a weakly-supervised contrastive loss and graph topology augmentation function to further improve model performance during the self-training process.













\section{Proposed Approach}

We start by introducing the notations used throughout this paper. We let bold uppercase letters represent matrices and bold lowercase letters denote vectors. Let $G = (\mathcal{V}, \mathcal{E}, \mathbf{X})$ denote an undirected graph with nodes $\mathcal{V}$ and edges $\mathcal{E}$. Let $n$ denote the number of nodes and $m$ the number of edges. The nodes in $G$ are described by the attribute matrix $\mathbf{X} \in \mathbb{R}^{n \times f}$, where $f$ denotes the number of features per node. The graph structure of $G$ is described by the adjacency matrix $\mathbf{A} \in \{0, 1\}^{n \times n}$, while $\Tilde{\mathbf{A}}$ stands for the adjacency matrix for a graph with added self-loops. We let $\Tilde{\mathbf{D}}$ be the diagonal degree matrix of $\Tilde{\mathbf{A}}$ and $\mathbf{S} = \Tilde{\mathbf{D}}^{-\frac{1}{2}} \Tilde{\mathbf{A}}\Tilde{\mathbf{D}}^{-\frac{1}{2}}$ denote the symmetric normalized adjacency matrix with self-loops. The class (or label) matrix is represented by $\mathbf{Y} \in \mathbb{R}^{n \times c}$, where $c$ denotes the number of classes.

\smallskip
\noindent\textbf{Problem Definition.}  Given an input graph $G = (\mathcal{V}, \mathcal{E}, \mathbf{X})$, where the node set $\mathcal{V}$ is divided to two disjoint node sets $\mathcal{V}^L$ and $\mathcal{V}^U$. In this paper we focus on the semi-supervised node classification task under the limited labeled data setting. Specifically, suppose that the labels of the labeled training set $\mathcal{V}^L$ are given, where each node class in $\mathcal{V}^L$ only have few labeled nodes (could be either balanced or imbalanced for different classes), the goal is to predict the labels of the unlabeled nodes in $\mathcal{V}^U$. Note that if each class has the same number of $K$-labeled nodes, the studied problem can be called few-shot semi-supervised node classification.

\smallskip
\noindent\textbf{Architecture Overview.} In this section, we propose an augmented graph self-training framework (AGST) to solve the problem of semi-supervised node classification with only a few labeled nodes. Compared to existing efforts, AGST is able to address the structural and semantic bottlenecks under the limited labeled data setting by virtue of two focal designs: (1) a new GST backbone with a graph topology augmentation function that is able to leverage long-range node interactions while alleviating the structure noise; and (2) a weakly-supervised contrastive loss that enhances the semantic structures of the input graph by aligning the semantic similarities between pseudo-labeled data and gold-labeled data. A detailed illustration of the proposed approach is provided in Figure \ref{fig:framework}.

\subsection{Augmenting Structural Knowledge in GST}

For semi-supervised node classification, graph neural predictors commonly have a large variance and easy to overfit when the labeled training data is extremely limited~\cite{li2018deeper,sun2020multi}. Though previous GST methods partially alleviate this issue by expanding the labeled training set, they still suffer from the incapability of leveraging long-range node interactions and handling structure noise in nature: on the one hand, if the teacher and student model shares the same shallow GNN architecture, the over-squashing issue will largely impede the effective propagation of feature-label patterns when multiple layers are deployed; on the other hand, the missing or noisy edges in the input graph may also
distort the information flow. To better exploit the useful graph structural knowledge, we go beyond the existing GST architectures and develop a decoupled GST backbone with a structural data augmentation module.



\smallskip
\noindent\textbf{Teacher Model.} In a self-training framework, the teacher model serves the role of generating pseudo labels on unlabeled data to augment the limited training set. The teacher model in our decoupled GST framework is a Label Propagation (LP) module that enables long-range propagation of label signals for computing the pseudo labels. This way the pseudo labels preserve both local and global structure knowledge when further training the student model. Specifically, the objective of LP is to find a prediction matrix $\mathbf{\hat{Y}} \in \mathbb{R}^{n \times c}$ that agrees with the label matrix $\mathbf{Y}$ while being smooth on the graph such that nearby vertices have similar soft labels:
\begin{equation}
    \mathbf{\hat{Y}} = \arg \min_{\mathbf{\hat{Y}}} \Big( \underbrace{\mathbf{Tr}(\mathbf{\hat{Y}}^{\mathrm{T}} (\mathbf{I} - \mathbf{S} )\mathbf{\hat{Y}})}_{\text{smoothness constraint}} + \ \underbrace{\mu ||\mathbf{\hat{Y}} - \mathbf{Y} ||_2^2}_{\text{fitting constraint}} \Big),
\end{equation}
where $\mu$ is a positive parameter balances the trade-off between these two competing constraints. The smoothness term smooths each column of the prediction matrix along the graph structure, while the fitting term enforces the prediction matrix $\mathbf{\hat{Y}}$ to agree with the label matrix $\mathbf{Y}$.

By solving the above unconstrained optimization function, a closed-form solution can be computed as $\mathbf{\hat{Y}} = (1 - \alpha) (\mathbf{I} - \alpha \mathbf{S} )^{-1} \mathbf{Y}$, where $\alpha  = \frac{1}{\mu + 1}$. As derived by Zhou et al.~\cite{zhou2004learning}, the solution can be approximated via the following iteration:
\begin{equation}
    \mathbf{Y}^{(t+1)} = \alpha \mathbf{S} \mathbf{Y}^{(t)} + (1 - \alpha)  \mathbf{Y}^{(0)},
    \label{eq:lp}
\end{equation}
where $\mathbf{Y}^{(0)} = \mathbf{Y}$, which converges to $\mathbf{\hat{Y}}$ rapidly. Here $1 - \alpha$ can be naturally connected with the teleport probability in Personalized PageRank~\cite{klicpera2019predict}. With an appropriate $\alpha$, the smoothed labels can avoid losing the focus on local neighborhood~\cite{li2018deeper} even using infinitely many propagation steps.


\smallskip
\noindent\textbf{Student Model.} Since the LP-based teacher model supports long-range propagation without losing its focus on the local neighborhood, both local and global structure knowledge will be captured in the computed soft pseudo labels. Next, we develop a Feature Transformation (FT) module as the student model to distill the knowledge from the teacher model and meanwhile learn the feature knowledge by transforming the node features to class labels. The student model is composed of an encoder network $f_{\bm\theta}(\cdot)$ followed by a prediction network $g_{\bm\phi}(\cdot)$. For node $v_i$, the class prediction can be computed:
\begin{equation}
\hat{\mathbf{p}}_i = g_{\bm\phi} (\mathbf{z}_i), \quad \mathbf{z}_i = f_{\bm\theta}(\mathbf{x}_i), 
\end{equation}
where the predicted label $\hat{\mathbf{p}}_i$ is computed based on the node features $\mathbf{x}_i$. Specifically, the encoder network $f_{\bm\theta}(\cdot)$ is built with a 2-layer MLP and the prediction network $ g_{\bm\phi}(\cdot)$ is a feed-forward layer followed by softmax, producing a vector of confidence scores.

To learn the student model, instead of using hard pseudo labels as previous GST methods, here we consider the soft pseudo labels generated by the LP-based teacher model as the ground-truth, and compute the standard cross-entropy loss on unlabeled nodes:
\begin{equation}
    \mathcal{L}_{\text{CE}}^U = - \sum_{v_i \in \mathcal{V}^U} \sum_{c=1}^{C} \hat{y}_{i}^c \log \hat{p}_{i}^c.
    \label{eq:cls_unlabeled}
\end{equation}

In addition, we derive another cross-entropy loss for the nodes from the labeled training set:
\begin{equation}
    \mathcal{L}_{\text{CE}}^L = - \sum_{v_i \in \mathcal{V}^L} \sum_{c=1}^{C} y_{i}^c \log \hat{p}_{i}^c.
    \label{eq:cls_labeled}
\end{equation}

By jointly optimizing the above losses, we are able to learn a simple yet effective student model for semi-supervised node classification with only few labels.




\smallskip
\noindent\textbf{Graph Topology Augmentation.} Real-world graphs commonly come with a certain level of structure noise, which could be induced by either partial observation, graph pre-processing, or even adversarial attacks~\cite{chen2020measuring,zhu2021deep,zhao2021data}. Since labeled nodes are extremely limited, the feature patterns of labeled nodes will be even harder to propagate to unlabeled nodes due to the imperfect graph structure. Considering that message-passing is a type of Laplacian Smoothing~\cite{klicpera2019predict}, representations of nodes belonging to different classes will become inseparable due to the existence of many unnecessary inter-class edges. As such, we argue that it is helpful to eliminate potentially noisy edges and strengthen the connections between similar nodes for better preserving the graph structure knowledge and improve the effectiveness of message-passing. 





To this end, after the student model converges in each self-training iteration, we in turn use it to refine the graph topology by strengthening intra-class edges and reducing inter-class connections. Specifically, given the predicted label matrix $\hat{\mathbf{P}}$, the edge probability matrix $\hat{\mathbf{A}}$ (symmetric) is computed by:
\begin{equation}
    \hat{\mathbf{A}} = \sigma (\hat{\mathbf{P}}^\mathrm{T} \hat{\mathbf{P}}), \quad \hat{\mathbf{P}} =  g_{\bm\phi}(f_{\bm\theta}(\mathbf{X})),
    \label{eq:edge_prob}
\end{equation}
where $\hat{\mathbf{A}}_{ij}$ denotes the probability that node $v_i$ and $v_j$ belong to a same class, $\sigma$ is an element-wise sigmoid function. 

Based on the Homophily principle~\cite{mcpherson2001birds} that assumes similar nodes are likely to be connected, thus we add/remove the edge $e_{ij}$ in the original adjacency matrix $\mathbf{A}$ if the edge probability $\hat{\mathbf{A}}_{ij}$ is larger/less than a threshold. Specifically, we add top $\beta_a |\mathcal{E}|$ non-exist (intra-class) edges with highest edge
probabilities, and removes the $\beta_r |\mathcal{E}|$ existing (inter-class) edges with lowest edge probabilities, where $\beta_a, \beta_r \in [0,1]$.



\smallskip
\noindent\textbf{Design Discussion.} It is noteworthy that such a design has the following unique advantages: (1) it not only enables long-range propagation of feature-label patterns by decoupling the transformation and propagation steps, but also improve the propagation process by using the learned student model to augment the input graph structure; (2) different from standard self-training paradigm where teacher and student have the same architecture, our decoupled backbone uses a LP module as the teacher and a MLP as student, which is parameter-less and evidently efficient; and (3) previous GST methods need to select unlabeled samples with high confidence as training targets. However, many of these selected predictions are incorrect due to the poor calibration of neural networks~\cite{rizve2021defense}. Our approach uses propagated soft pseudo labels to circumvent the process of pseudo label selection.

\subsection{Augmenting Semantic Knowledge in GST}

Despite the effectiveness of the above design, the generated pseudo labels could introduce complex feature patterns and noisy training labels since the teacher model is trained with few labels, which exacerbates the difficulty of learning well-separated decision boundaries. Hence, how to enforce pseudo-labeled nodes to have aligned usage of gold-labeled ones is another important factor for improving the semantic knowledge of GST under the low-data regime.






To this end, we propose a \textit{weakly-supervised contrastive loss} that mitigates pseudo label noise and enhances semantic structure learning. Specifically, a contrastive loss~\cite{chen2020simple} encourages the similarity function to assign large values to the positive pairs and small values to the negative pairs. With similarity measured by dot product, a form of a contrastive loss function, called InfoNCE~\cite{oord2018representation,he2020momentum} has been widely used in self-supervised learning:
\begin{equation}
    \mathcal{L}_{\text{InfoNCE}} = \sum_{i=1}^n - \log \frac{\exp( \mathbf{z}_i \cdot \mathbf{z}_i'/\tau)}{\sum_{j=0}^r\exp(\mathbf{z}_i \cdot \mathbf{z}_j'/\tau)},
\end{equation}
where $\mathbf{z}_i'$ are positive embedding for $\mathbf{z}_i$, and $\mathbf{z}_j'$ includes one positive embedding and $r$ negative embedding for other instances. Here $\tau$ is a temperature hyper-parameter. 


Different from the unsupervised contrastive  loss~\cite{oord2018representation} which only preserves the local smoothness around each instance, our goal is to improve the utility of pseudo-labeled nodes to learn better semantic structures of the input graph. To achieve this, we first compute the similarity distribution between pseudo-labeled samples and different class prototypes in the feature space:
\begin{equation}
    s_i^j = \frac{\exp(\mathbf{z}_i \cdot \mathbf{c}_{j}/\tau)}{\sum_{c=1}^C\exp(\mathbf{z}_i \cdot \mathbf{c}_c/\tau)}, \quad \mathbf{c}_c = \frac{1}{|\mathcal{V}^L_c|} \sum_{v_i \in \mathcal{V}^L_c} \mathbf{z}_i,
\end{equation}
where $\mathcal{V}^L_c$ denotes the labeled node set of class $c$ and $\mathbf{c}_c$ is the corresponding class prototype computed as the average of the labeled examples in class $c$. 

Due to the existence of incorrect pseudo labels, there could be an inconsistency between the latent feature space and the pseudo label space. Here we apply the following rule to obtain a filtered set of pseudo-labeled nodes $\mathcal{V}^{\widehat{U}}$:
\begin{equation}
  s^{\hat{y}_i}_i > 1/C, \quad \hat{y}_i = \arg \max_j
\hat{y}_i^j,
\end{equation}
where $\hat{y}_i$ is the hard pseudo label of node $v_i$ with the maximum confidence score. For each pseudo-labeled node $v_i$, we consider it trustworthy if its embedding similarity (to its pseudo-labeled prototype) $s^{\hat{y}_i}_i$ is higher than uniform probability, this way we largely reduce the risk of noisy training on  hard pseudo labels.


With the calibrated pseudo-labeled node set, next we try to enhance the intra-class compactness and inter-class separability of learned node representations. Specifically, our contrastive loss encourages each node clustering around their corresponding class prototypes, which can be formulated as follows:
\begin{equation}
    \mathcal{L}_{\text{CL}} = \sum_{v_i \in \mathcal{V}^{\widehat{U}}} - \log \frac{\exp(\mathbf{z}_i \cdot \mathbf{c}_{\hat{y}_i}/\tau)}{\sum_{c=1}^C\exp(\mathbf{z}_i \cdot \mathbf{c}_c/\tau)},
    \label{eq:contrastive}
\end{equation}
where $\mathbf{c}_{\hat{y}_i}$ denotes the corresponding prototype of node $v_i$. For any pseudo-labeled node $v_i$, its embedding, $\mathbf{z}_i$, is treated as the anchor, the embedding of its corresponding class prototype $\mathbf{c}_{\hat{y}_i}$ forms the positive sample, and the embeddings of other class prototypes are naturally
regarded as negative samples. This loss will be optimized to reduce the variance of pseudo-labeled nodes that share the same semantic, while pushing away instances from different classes. 

For the sake of stable training, we follow the idea of MoCo~\cite{he2020momentum} and use the node representations learned from a \textit{momentum encoder} parameterized by $f_{\bm\theta'}(\mathbf{x}_i)$ to compute the momentum prototype of each class. Note that the momentum encoder has the same architecture as the encoder network, and its parameters are the moving-average of the encoder's parameters. Formally:
\begin{equation}
    \bm\theta_t' = m \cdot \bm\theta_{t-1}' + (1 - m) \cdot \bm\theta_{t}
\end{equation}
where $m$, $\bm\theta$, and $\bm\theta'$ are momentum, encoder parameters, and the momentum encoder parameters, respectively.




\begin{algorithm}[t!]
\caption{The learning algorithm of AGST.}
\label{alg:AGST}
\LinesNumbered
\small
\KwIn{The input graph $G = (\mathcal{V}, \mathcal{E})$ with labeled node set $\mathcal{V}^L$ and unlabeled node set $\mathcal{V}^U$, self-training iterations $I$}
\KwOut{The well-trained student model}

Initialize the parameters $\bm\theta$ and $\bm\phi$

\For{$i = 1, 2, \dots, I$}{
    \LeftComment{\textit{Label Propagation (Teacher)}}\\
    Generate soft pseudo labels on unlabeled nodes by Eq. (\ref{eq:lp}); 

    \LeftComment{\textit{Feature Transformation (Student)}}\\
    \While{not converge}{
        Compute the classification loss according to Eq. (\ref{eq:cls_unlabeled}), Eq. (\ref{eq:cls_labeled});\\
        Compute the contrastive loss according to Eq. (\ref{eq:contrastive});\\
        Update the student model's parameters by optimizing the joint loss in Eq. (\ref{eq:joint_loss});\\
    }
    
    \LeftComment{\textit{Graph Topology Augmentation}}\\
    Compute the edge probability matrix using Eq. (\ref{eq:edge_prob});\\
    Augment the input graph by adding/removing edges  \\
     
}
    
\Return The student model

\end{algorithm}

\subsection{Model Training}
Given the above focal designs, we train the AGST framework with an iterative learning fashion. In each self-training iteration, the teacher model first generate pseudo labels and we then optimize the student model until converge. Afterwards, the graph structure will be refined by the topology augmentation function and fed to the next iteration.

To train the student model end-to-end, we jointly optimize the classification losses and the weakly-supervised contrastive loss. The full training objective is defined as follows:
\begin{equation}
    \mathcal{L} =  \mathcal{L}_{\text{CE}}^L + \lambda_1 \mathcal{L}_{\text{CE}}^U + \lambda_2 \mathcal{L}_{\text{CL}},
    \label{eq:joint_loss}
\end{equation}
where $\lambda_1$ and $\lambda_2$ are balancing parameters. According to our preliminary experiments, simply setting the parameters $\lambda_1$ and $\lambda_2$ as $1$ and $0.1$ can offer stable and strong performance in practice. By minimizing the training objective, the rich unlabeled data and the scarce yet valuable labeled data work collaboratively to provide additional supervision signals for learning discriminative prediction model. The detailed learning process of AGST is presented in Algorithm \ref{alg:AGST}.
Note that in each iteration, we refine the graph topology based on the original graph structure instead of the previously refined graph, since informative edges might be accidentally removed at the early stage of the training procedure. To reduce computational complexity, we only consider the edge between node $v_i$ and $v_j$ as candidate only when they have the same hard labels when adding edges. Further, by refining the graph structure, it is essential to repeat training both the teacher and student models, which connects naturally to the iteration loops in conventional self-training.



\section{Experiments}
In this section, we start by introducing the setup of our experiments. Then we conduct experiments on benchmark datasets to show the effectiveness of the proposed framework. The implementation details and additional results can be found in Appendix.

\begin{table*}[t!]
\centering
\caption{Test accuracy of semi-supervised node classification with few labels (balanced training setting): mean accuracy ($\%$) with 95$\%$ confidence interval.}


\scalebox{0.91}{
\begin{tabular}{@{}lcccccccccccccccccc@{}}
\toprule
\rule{0pt}{10pt} \multirow{2}{*}{\textbf{Method}}  & \multicolumn{3}{c}{Cora}  & &  \multicolumn{3}{c}{CiteSeer}  & & \multicolumn{3}{c}{PubMed} \\ \cline{2-4} \cline{6-8} \cline{10-12} 

\rule{0pt}{10pt} & \multicolumn{1}{c}{3-shot}  & \multicolumn{1}{c}{5-shot}  & \multicolumn{1}{c}{10-shot} & & \multicolumn{1}{c}{3-shot} &  \multicolumn{1}{c}{5-shot} & \multicolumn{1}{c}{10-shot} & &
\multicolumn{1}{c}{3-shot} & \multicolumn{1}{c}{5-shot} & \multicolumn{1}{c}{10-shot}

\\ \midrule

LP        & $52.76\pm0.92$   & $58.72\pm0.79$   & $64.03\pm0.65$ & & $34.87\pm0.93$ & $37.58\pm0.81$ & $41.74\pm0.50$ & & $59.58\pm0.98$ & $62.32\pm0.94$ & $67.02\pm0.75$ \\

GCN    & $56.31\pm0.81$   & $64.18\pm0.66$ & $72.87\pm0.53$  & & $47.59\pm0.90$ & $54.27\pm0.81$ & $62.26\pm0.57$ & & $59.24\pm0.81$ & $66.40\pm0.85$ & $72.37\pm0.74$  \\

GAT      & $63.39\pm0.98$   & $69.93\pm0.84$  & $76.44\pm0.35$ & & $51.62\pm0.97$ & $58.67\pm0.81$ & $65.13\pm0.51$ & & $64.72\pm0.91$ & $68.32\pm0.90$ & $73.85\pm0.60$  \\

SGC    & $55.94\pm0.97$   & $59.77\pm0.97$ & $67.76\pm0.91$  & & $52.60\pm0.92$ & $58.94\pm0.85$ & $64.92\pm0.54$ & & $58.74\pm0.92$ & $64.72\pm0.91$ & $69.02\pm0.83$  \\
\midrule

GLP   & $65.99\pm0.94$   & $72.31\pm0.89$  & $77.56\pm0.43$ & & $50.46\pm0.96$ & $59.09\pm0.88$ & $66.06\pm0.38$ & & $66.31\pm0.95$ & $72.59\pm0.73$ & $75.82\pm0.58$  \\
IGCN   & $66.91\pm0.91$   & $72.78\pm0.85$ & $78.27\pm0.31$  & & $50.99\pm0.97$ & $59.53\pm0.89$ & $66.51\pm0.39$ & & $66.23\pm0.97$ & $71.96\pm0.85$ & $75.97\pm0.50$  \\

CGPN  &  \underline{$71.88\pm2.52$}  & $71.83\pm3.14$ & $74.85\pm1.54$  & & $\mathbf{62.54\pm3.56}$ & \underline{$62.20\pm1.63$} & $63.76\pm1.09$ & & \underline{$68.21\pm3.89$} & $71.21\pm2.90$ & $75.44\pm2.53$  \\
\midrule
PTA   & $69.21\pm0.99$   & \underline{$73.98\pm0.73$}  & \underline{$78.69\pm0.39$} & & $54.18\pm0.94$ & $61.13\pm0.86$ & $66.69\pm0.48$ & & $67.69\pm0.92$ & $72.28\pm0.82$ & \underline{$76.47\pm0.51$} \\

ST-GCNs   & $65.85\pm0.94$   & $71.16\pm0.87$  & $76.54\pm0.49$ & & $49.85\pm0.95$ & $61.39\pm0.91$ & \underline{$68.58\pm0.36$} & & $65.99\pm0.93$ & $70.26\pm0.98$ & $74.10\pm0.63$  \\
M3S  & $64.01\pm0.71$   & $69.26\pm0.75$ & $77.20\pm0.41$  & & $50.31\pm0.88$ & $59.72\pm0.82$ & $65.99\pm0.41$ & & $66.01\pm0.90$ & \underline{$72.38\pm0.85$}  & $75.31\pm0.49$  \\

\textbf{AGST} & $\mathbf{73.20\pm0.79}$  &  $\mathbf{76.95 \pm0.72}$ & $\mathbf{80.89\pm0.42}$ && $\underline{60.31\pm0.85}$ & $\mathbf{66.58\pm0.80}$ & $\mathbf{72.30\pm0.39}$ && $\mathbf{71.16\pm0.93}$ & $\mathbf{74.87\pm0.89}$ & $\mathbf{77.84\pm0.41}$\\

\bottomrule
\end{tabular}}

\medskip

\scalebox{0.91}{
\begin{tabular}{@{}lcccccccccccccccccc@{}}
\toprule
\rule{0pt}{10pt} \multirow{2}{*}{\textbf{Method}}  & \multicolumn{3}{c}{Coauthor-CS}  & &  \multicolumn{3}{c}{Coauthor-Physics}  & & \multicolumn{3}{c}{Amazon-Photo} \\ \cline{2-4} \cline{6-8} \cline{10-12} 

\rule{0pt}{10pt} & \multicolumn{1}{c}{3-shot}  & \multicolumn{1}{c}{5-shot}  & \multicolumn{1}{c}{10-shot} & & \multicolumn{1}{c}{3-shot} &  \multicolumn{1}{c}{5-shot} & \multicolumn{1}{c}{10-shot} & &
\multicolumn{1}{c}{3-shot} & \multicolumn{1}{c}{5-shot} & \multicolumn{1}{c}{10-shot}

\\ \midrule

LP   & $57.77\pm0.77$ & $62.09\pm0.60$    & $66.18\pm0.36$ & & $73.46\pm0.93$ & $76.94\pm0.61$ & $80.55\pm0.41$ & & $69.24\pm0.92$ & $73.43\pm0.72$ & $77.78\pm0.61$ \\

GCN  & $77.17\pm0.79$ & $84.09\pm0.59$   & $89.01\pm0.98$ & & $82.49\pm0.88$ & $87.50\pm0.69$ & $90.78\pm0.38$ & & $69.54\pm0.99$ & $74.42\pm0.97$ & $80.30\pm0.78$ \\

GAT   & $79.66\pm0.75$ & $85.11\pm0.49$  & $89.34\pm0.19$ & & $86.07\pm1.16$ & $89.35\pm0.48$ & $91.64\pm0.48$ & & $70.47\pm1.19$ & $77.89\pm1.05$ & $82.39\pm1.11$ \\

SGC   & $84.93\pm0.57$ & $88.11\pm0.35$  & $90.13\pm0.99$ & & $87.55\pm0.64$ & $87.68\pm0.39$ & $91.38\pm0.31$ & & $75.05\pm0.88$ & $78.73\pm0.69$ & $84.14\pm0.45$ \\
\midrule

GLP  & $84.58\pm0.61$ & $87.36\pm0.61$ & $91.59\pm0.15$ & & $89.34\pm0.99$ & $91.52\pm0.32$ & \underline{$93.02\pm0.20$} & & $75.11\pm1.19$ & $81.99\pm0.97$ & $85.33\pm0.38$ \\
IGCN  & $84.26\pm0.47$ & $86.45\pm0.33$  & $90.82\pm0.13$ & & $89.82\pm0.57$ & $91.33\pm0.29$ & $92.78\pm0.21$ & & $75.36\pm0.98$ & $82.10\pm0.89$ & $85.50\pm0.32$ \\
CGPN & \underline{$88.96\pm3.37$} & $89.14\pm3.27$  & $90.37\pm2.14$ & & \underline{$90.06\pm3.48$} & \underline{$91.76\pm2.33$} & $92.56\pm2.22$ & & \underline{$83.57\pm3.24$} & \underline{$84.74\pm2.63$} & \underline{$87.78\pm2.44$} \\
\midrule

PTA   & $86.56\pm0.46$ & $89.43\pm0.31$  & $90.72\pm0.18$ & & $88.62\pm0.60$ & $90.36\pm0.53$ & $92.15\pm0.32$ & & $77.43\pm0.89$ & $82.63\pm0.76$ & $85.51\pm0.74$ \\

ST-GCNs  & $88.34\pm0.46$ & \underline{$89.68\pm0.45$}  & \underline{$91.39\pm0.14$} & & $87.61\pm0.69$ & $90.23\pm0.39$ & $91.75\pm0.21$ & & $73.86\pm1.53$ & $81.93\pm1.09$ & $85.54\pm0.67$ \\
M3S & $84.11\pm0.46$ & $86.96\pm0.41$  & $91.08\pm0.11$ & & $89.12\pm0.55$ & $91.27\pm0.31$ & $92.93\pm0.25$ & & $74.96\pm0.97$ & $81.88\pm0.93$ & $85.42\pm0.37$ \\

\textbf{AGST}  & $\mathbf{90.29\pm0.33}$ & $\mathbf{91.31\pm0.37}$  & $\mathbf{92.76\pm0.12}$ & & $\mathbf{92.86\pm0.62}$ & $\mathbf{93.04\pm0.47}$ & $\mathbf{94.37\pm0.24}$ & & $\mathbf{85.08\pm0.89}$ & $\mathbf{86.53\pm0.92}$ & $\mathbf{89.27\pm0.62}$ \\

\bottomrule
\end{tabular}}

\label{table:imbalanced}
\end{table*}

\subsection{Experimental Setup}
\label{sec:experiment}






\smallskip
\noindent \textbf{Evaluation Datasets.} 
We adopt six graph benchmark datasets to demonstrate the effectiveness of the proposed approach for semi-supervised node classification. Specifically, \textbf{Cora}~\cite{sen2008collective}, \textbf{CiteSeer}~\cite{sen2008collective}, and \textbf{PubMed}~\cite{namata2012query} are three most widely used citation networks. \textbf{Coauthor-CS}~\cite{shchur2018pitfalls} and \textbf{Coauthor-Physics}~\cite{shchur2018pitfalls} are two co-authorship graphs based on the Microsoft Academic Graph. \textbf{Amazon-Photo}~\cite{shchur2018pitfalls} is an Amazon product co-purchase networks. 



To provide a robust and fair comparison between different models on each dataset, we conduct evaluation under two low-data settings with different data splitting protocols as follows:
\begin{itemize}[leftmargin=*,noitemsep,topsep=1.5pt]
    \item \textit{Balanced training setting.} Similar to the setting in \cite{shchur2018pitfalls,liu2020towards}, for each dataset, we sample few (i.e., $K$-shot) labeled nodes per class as the training set, $30$ nodes per class as the validation set, and the rest as the test set. We conduct 100 runs for random training/validation/test splits to ensure a fair comparison.
    
    \item \textit{Imbalanced training setting.} In this setting, we strictly follow the setup in \cite{li2018deeper,sun2020multi} and randomly split the data into one small sample subset for training, and the test sample subset with 1000 samples. Following this line of work, we report the mean accuracy of 10 runs without validation to make fair comparison. 
    

\end{itemize}






\smallskip
\noindent \textbf{Compared Methods.} In our experiments, we compare the proposed approach AGST with both classic and state-of-the-art methods on the semi-supervised node classification task. In addition to the traditional semi-supervised learning method Label Propagatio (LP), other baseline methods can be generally categorized into three classes: (1) \textit{Vanilla GNNs} that only allows shallow message passing, including GCN~\cite{kipf2017semi}, GAT~\cite{velickovic2018graph}, SGC~\cite{wu2019simplifying}; (2) \textit{Data-efficient GNNs} that can better propagate messages from limited labeled data, including GLP~\cite{li2019label}, IGCN~\cite{li2019label} and CGPN~\cite{wan2021contrastive}; (3) \textit{Self-training GNNs} that adopt the teacher-student architecture to leverage pseudo labels during training, including PTA~\cite{dong2021equivalence}, ST-GCNs~\cite{li2018deeper} (and its variants), and M3S~\cite{sun2020multi}.

\begin{table*}[t!]
\caption{Test accuracy of semi-supervised node classification with few labels (imbalanced training setting): mean accuracy ($\%$) with 95$\%$ confidence interval. Results of baseline methods are borrowed from M3S~\cite{sun2020multi}.}
\centering
\scalebox{0.925}{
\begin{tabular}{@{}c|c|cc|ccccc|cc@{}}
\toprule

\textbf{Dataset} & Label Rate&  LP & GCN  & Co-train & Self-train & Union & InterSection & M3S & \textbf{AGST} (teacher) & \textbf{AGST} (student) \\ \midrule
\multirow{4}{*}{Cora} & 0.5\% & 57.6 &  50.6 & 53.9 & 56.8 & 55.3 & 50.6 & \underline{61.5} & $60.2\pm1.02$ & $\mathbf{70.2\pm0.93}$ \\
& 1.0\% & 61.0 & 58.4  & 57.0 & 60.4 & 60.0 & 60.4 & \underline{67.2} & $63.6\pm0.87$ & $\mathbf{75.8\pm0.80}$\\
& 2.0\% & 63.5  & 70.0 & 69.7 & 71.7 & 71.7 & 70.0 & \underline{75.6} & $68.0\pm0.48$ & $\mathbf{78.3\pm0.51}$\\
  & 3.0\% & 64.3 & 75.7 & 74.8 & 76.8 & 77.0 & 74.6 & \underline{77.8} & $70.2\pm0.40$ & $\mathbf{80.1\pm0.36}$\\
\midrule
\multirow{4}{*}{CiteSeer} & 0.5\% & 37.7 &  44.8 & 42.0 & 51.4 & 48.5 & 51.3 & \underline{56.1} & $46.8\pm0.91$ & $\mathbf{63.3\pm0.88}$\\
& 1.0\% & 41.6 & 54.7  & 50.0 & 57.1 & 52.6 & 61.1 & \underline{62.1} & $54.2\pm0.82$ & $\mathbf{71.4\pm0.80}$\\
& 2.0\%  &  41.9 & 61.2  & 58.3 & 64.1 & 61.8 & 63.0 & \underline{66.4} & $59.5\pm0.52$ & $\mathbf{72.0\pm0.43}$\\
& 3.0\%   & 44.4 & 67.0 & 64.7 & 67.8 & 66.4 & 69.5 & \underline{70.3} & $61.6\pm0.39$ & $\mathbf{72.8\pm0.31}$\\

\midrule
\multirow{3}{*}{PubMed} & 0.03\% & 58.3 & 51.1 & 55.5 & 56.3 & 57.2 & 55.0 & \underline{59.5} & $59.3\pm1.25$ & $\mathbf{69.6\pm1.03}$\\
& 0.05\% & 61.3 & 58.0  & 61.6 & 63.6 & 64.3 & 58.2 & \underline{64.4} & $63.3\pm1.05$ & $\mathbf{73.5\pm0.98}$\\
& 0.1\%  &  63.8 & 67.5  & 67.8 & 70.0 & 70.0 & 67.0 & \underline{70.6} & $65.7\pm0.93$ & $\mathbf{78.9\pm0.89}$\\

\bottomrule
\end{tabular}}

\medskip

\label{table:balanced}
\end{table*}

\subsection{Main Results}
In our experiments, we evaluate the proposed framework AGST and all the baseline methods on semi-supervised node classification task in low-data settings, which aims to predict the missing node labels with only a few labeled nodes. 

\smallskip 
\noindent\textbf{Balanced Training Setting.}
We first compare the proposed framework AGST with baseline methods under the canonical few-shot semi-supervised setting, in which each node class is provided with few labeled samples. We run each model with 3, 5, 10 labeled nodes per class (i.e., 3-shot, 5-shot, and 10-shot) and report the average test accuracy under such balanced training setting in Table~\ref{table:imbalanced}. According to the reported results, we are able to make the following in-depth observations and analysis: 
 \begin{itemize}[leftmargin=*,noitemsep,topsep=1.5pt]
    \item Overall, AGST significantly outperforms all the baseline methods on each dataset based on the paired t-tests with $p < 0.05$. For example, AGST improves the best performing baseline model (i.e., PTA) on Cora obtains an $4.01\%$ improvement in 5-shot evaluation. This observation further proves that the design of AGST is effective for tackling the node classification problem when only few labels per class are given.

      \item Both deep GNNs and GST methods can achieve better performance over the shallow GNNs such as GCN and GAT when training data is extremely scarce. While compared to the deep GNNs, existing GST methods cannot achieve better performance in most cases, which verifies our claim that their shallow backbones largely restrict the effective propagation of label signals. Our framework AGST adopts a decoupled backbone that inherits the advantages of both deep GNN models and GST methods, which is more data-efficient.

     \item Though LP only relies on structure information, it can perform competitively with shallow GNNs on some datasets when training labels are extremely limited, such as Cora and PubMed. However, we also notice that LP become the worst performing method on datasets like CiteSeer and Coauthor CS. The main reason behind is that noisy graph structure could easily lead to incorrect propagation of label signals, which verifies the rationality and necessity of refining the graph topology in GST. 
     
     

 \end{itemize}
 

\smallskip 
\noindent\textbf{Imbalanced Training Setting.}
Furthermore, we follow the imbalanced training setting used in \cite{li2018deeper,sun2020multi} and conduct another set of experiments where each model is trained with different label rates, i.e., 0.5\%, 1\%, 2\%, 3\% on Cora and CiteSeer, 0.03\%, 0.05\%, 0.1\% on PubMed.  Compared to the balanced training setting, this evaluation setting is more challenging since the training labels for each class could varies a lot. We report the average accuracy on three datasets in Table \ref{table:balanced}. For a fair comparison, the results of baseline methods are 
borrowed from the previous work~\cite{sun2020multi}.
 \begin{itemize}[leftmargin=*,noitemsep,topsep=1.5pt]
     
     \item Similar to the balanced training setting, GCN that only uses limited receptive fields cannot achieve satisfactory results when labeled data is scarce and imbalanced. By incorporating pseudo labels into the learning process, methods including Co-train, Self-train, Union and Intersection are able to improve the performance of GCN with only few labels.
    
     \item  However, the performance of those baselines based on pseudo-labeling varies a lot under different datasets, which shows that it is practically difficult to select informative pseudo labels and mitigate the pseudo label noise. Though M3S partially address this by using the clustering methods, it still largely falls behind our approach due to the inability of leveraging large receptive fields and handling structure noise.

     \item Compared to the original LP, the teacher model in the proposed AGST framework achieves better performance, even though they both use the same label propagation algorithm. It demonstrates that the graph structure can be well refined by our graph topology augmentation function. Based on both labeled and unlabeled data, the student model further pushes forward the performance by performing feature transformation and weakly-supervised contrastive learning.  
     

 \end{itemize}

\smallskip 
\noindent\textbf{Standard (20-shot) Training Setting.} Next we examine the performance of AGST under the standard semi-supervised node classification tasks. Specifically, we randomly sample 20 labeled nodes for each class (i.e., 20-shot) as the training set and test the performance of different methods. According to the average performance reported in Table~\ref{table:20-shot}, we can observe that: (1) the performance gain of GST methods over the vanilla GNNs decreases since the standard training setting has more labeled data; (2) data-efficient GNN such as CGPN cannot perform well under the standard semi-supervised learning setting; (3) though AGST is mainly proposed for few-shot semi-supervised learning, it still achieves the best performance for the standard semi-supervised node classification task, illustrating the superiority of our approach.

\begin{table}[h]
\centering
\caption{Test accuracy on standard (20-shot) node classification: mean accuracy ($\%$) with 95$\%$ confidence interval.}
\scalebox{0.915}{
\begin{tabular}{@{}lccccccccc@{}}
\toprule

\rule{0pt}{10pt} \textbf{Method} & \multicolumn{1}{c}{Cora}  & \multicolumn{1}{c}{CiteSeer}  &  \multicolumn{1}{c}{PubMed} 
 &  \multicolumn{1}{c}{Coauthor-CS} 

\\ \midrule

LP        & $67.04\pm0.41$   & $45.29\pm0.34$   &  $69.78\pm0.54$
 & $72.24\pm0.24$ \\

GCN    & $77.85\pm0.33$   & $65.95\pm0.42$   &  $76.33\pm0.47$ & $90.92\pm0.11$ \\

GAT    & $76.85\pm0.34$   & $65.12\pm0.72$   &  $73.20\pm0.49$ 
 & $90.39\pm0.98$ \\

SGC    & $71.19\pm0.29$   & $69.20\pm0.37$   &  $72.13\pm0.66$ 
 & $91.03\pm0.21$\\
\midrule

GLP   & $79.33\pm0.27$   & $68.94\pm0.28$   &  $78.49\pm0.39$ 
 & $82.53\pm0.29$ \\
IGCN   & $80.11\pm0.31$   & $67.89\pm0.29$   &  $78.64\pm0.39$ 
 & $83.50\pm0.23$ \\
CGPN  & $74.12\pm1.54$    &  $67.34\pm1.07$  &  $75.81\pm1.26$
 & $89.92\pm1.10$ \\
 \midrule
 PTA  & \underline{$81.54\pm0.35$}   & $69.84\pm0.25$   &  \underline{$78.66\pm0.44$} 
 & \underline{$92.51\pm0.15$} \\
ST-GCNs   & $79.75\pm0.24$   & $70.26\pm0.23$   &  $78.12\pm0.30$ 
 & $91.61\pm0.11$ \\
M3S   & $78.11\pm0.39$   & \underline{$70.42\pm0.29$}   &  $77.98\pm0.29$ 
 & $91.90\pm0.18$ \\



\textbf{AGST}  & $\mathbf{82.57\pm0.22}$   & $\mathbf{71.52\pm0.11}$   &  $\mathbf{79.92\pm0.27}$ 
 & $\mathbf{93.27\pm0.13}$ \\

\bottomrule
\end{tabular}}
\label{table:20-shot}
\end{table}

\subsection{Ablation Study}

In this section, we further conduct ablation studies to demonstrate the contribution of each component in AGST and justify our architectural design choice. Here \textit{AGST-base} denotes the decoupled GST backbone of our framework. Meanwhile, we include another two variants by removing each of the other two key designs in the proposed framework. Specifically, \textit{w/o contrast} represent the variant of AGST that excludes the weakly-supervised contrastive loss, and \textit{w/o augment} is the variant without the graph topology augmentation function. Compared to the complete framework AGST, \textit{w/o contrast} loses part of the semantic knowledge and \textit{w/o augment} loses part of the 
structural knowledge.

\begin{figure}[h]

    \graphicspath{{figures/}}
    \centering
    \subfigure[\textbf{Cora}] 
    {
    \includegraphics[width=0.425\columnwidth]{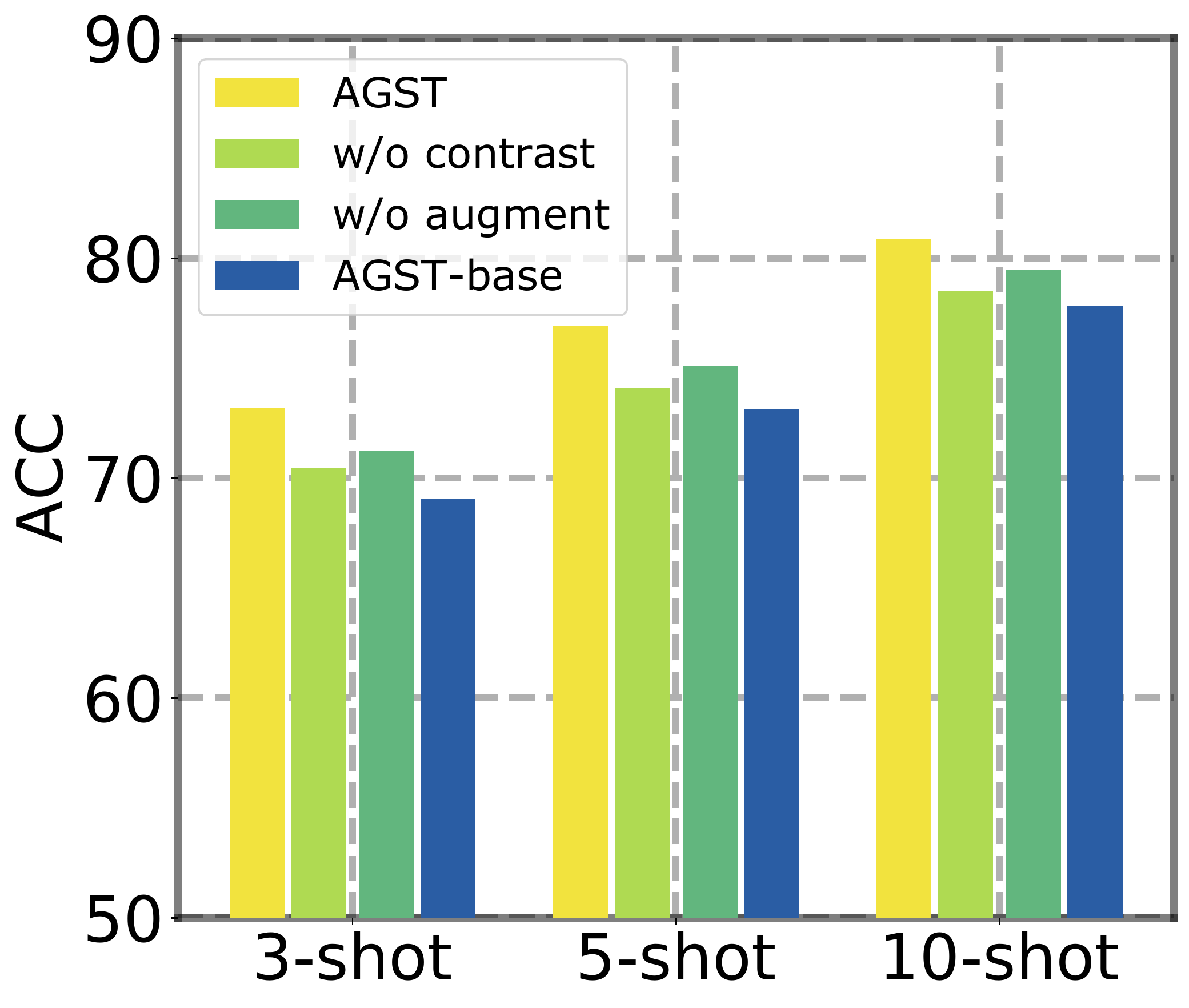}
    }
    \hspace{-0.1cm}
    \subfigure[\textbf{CiteSeer}]
    {
    \includegraphics[width=0.425\columnwidth]{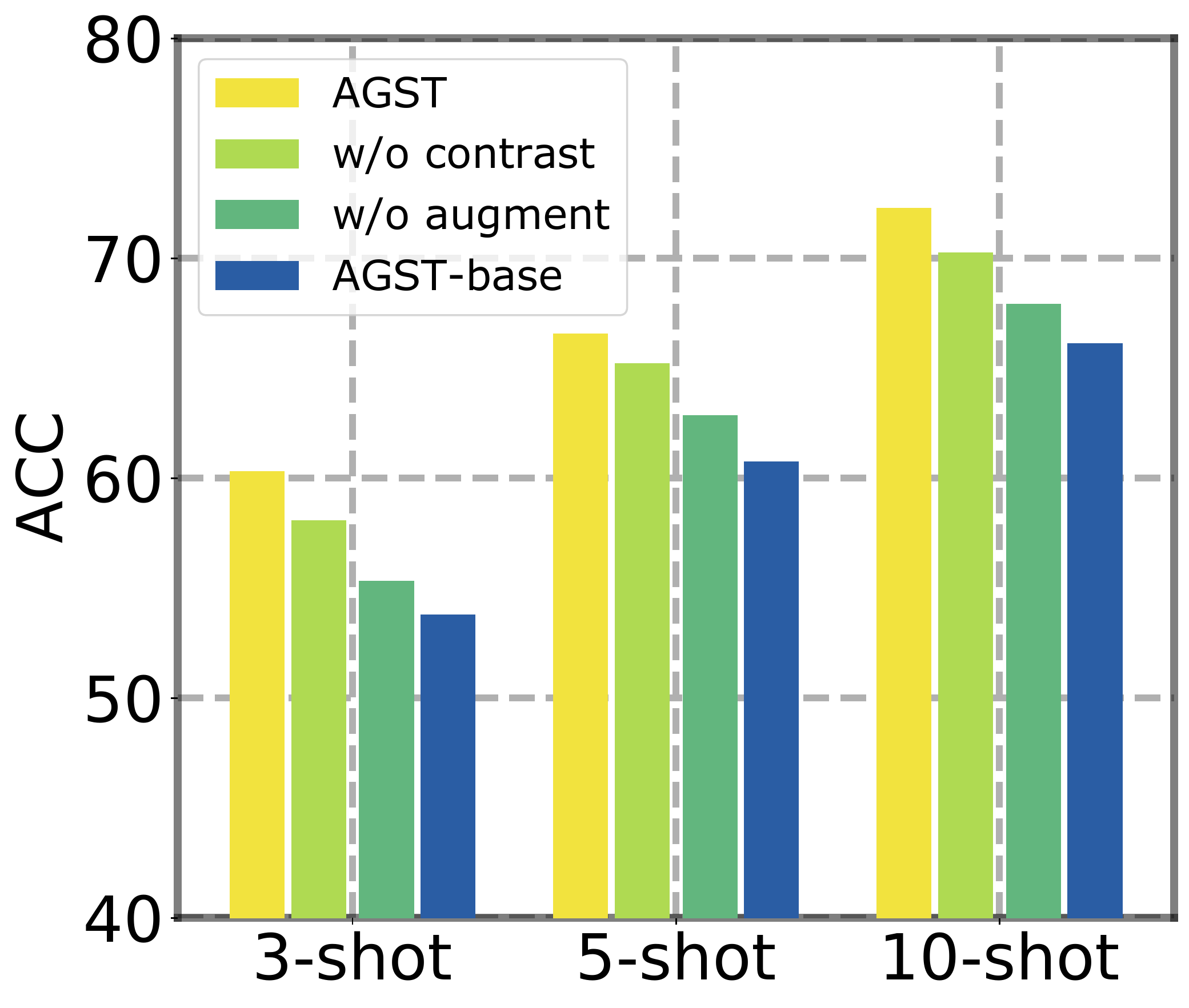}
    }
    \caption{Ablation results for different model variants.}%

    \label{fig:ablation}
\end{figure}

We report the accuracy results of each variant (balanced training) on two datasets Cora, CiteSeer in Figure \ref{fig:ablation}. It is apparent that the classification accuracy will decrease when any one of the focal components is removed, which reveals that both the weakly-supervised contrastive loss and the graph topology augmentation function make essential contributions to boosting the model performance. Meanwhile, compared to the conventional GNNs in Table ~\ref{table:balanced}, the backbone of AGST, i.e., \textit{AGST-base} is able to achieve better classification performance. Meanwhile, by comparing \textit{w/o augment} with \textit{AGST-base}, we can see that our contrastive loss brings further improvements. Notably, the importance of the topology augmentation function varies on different datasets, it usually have larger contribution on datasets with noisy graph structures, such as CiteSeer. 



\subsection{Parameter Analysis}
\begin{figure}[t]
    \graphicspath{{figures/}}
    \centering
    \subfigure[\textbf{Cora}]
    {
    \includegraphics[width=0.46\columnwidth]{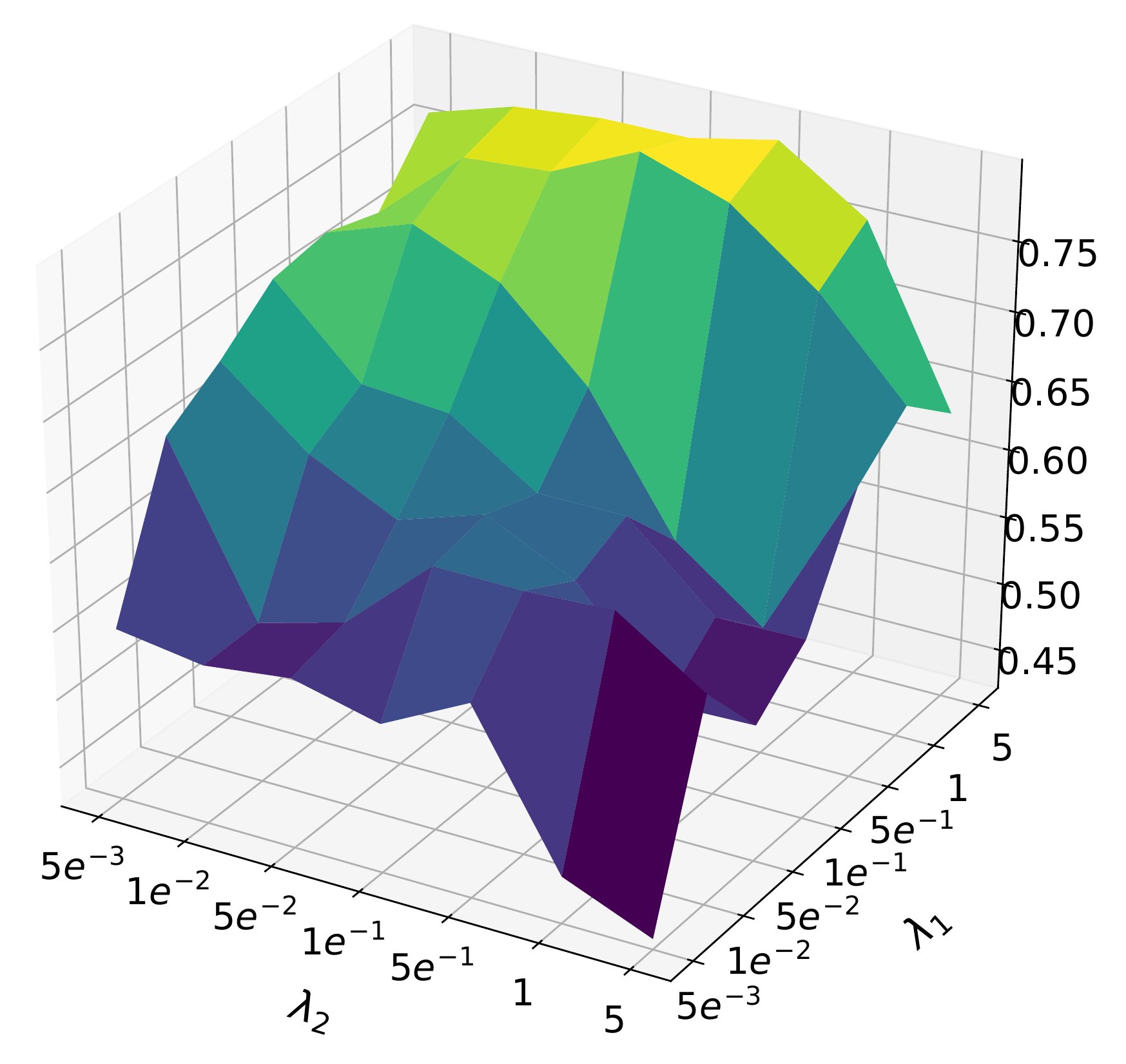}
    }
    \hspace{-0.1cm}
    \subfigure[\textbf{Citeseer}]
    {
    \includegraphics[width=0.45\columnwidth]{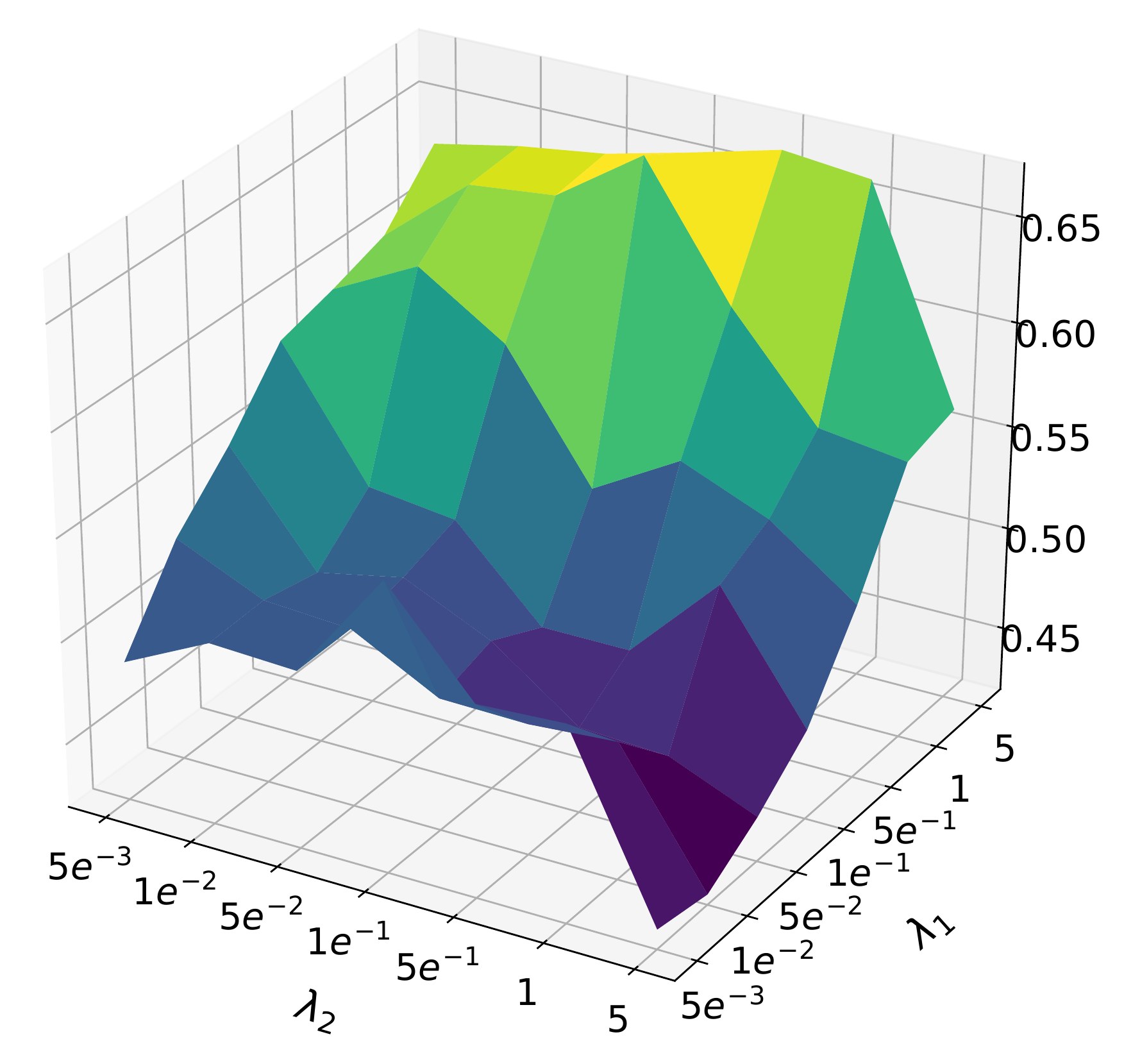}
    }
    \caption{Parameter analysis ($\lambda_1$, $\lambda_2$) results on Cora, Citeseer (5-shot).}%
    \label{fig:parameter_3d}
\end{figure}

\smallskip
\noindent\textbf{Loss Balancing Parameters.} We explore the sensitivity of the model performance in terms of two important hyper-parameters $\lambda_1$ and $\lambda_2$, which are in the final objective function of
AGST. $\lambda_1$ controls the contribution of the cross-entropy loss on pseudo labels (i.e.,  $\mathcal{L}_{\text{CE}}^U$) and $\lambda_2$ controls the contribution of weakly-supervised contrastive loss (i.e., $ \mathcal{L}_{\text{CL}}$). Specifically, we vary the values of $\lambda_1$ and $\lambda_2$ as $\{0.005, 0.01, 0.05, 0.1, 0.5, 1, 5\}$ on the Cora and Citeseer datasets and report the results of AGST in Figure \ref{fig:parameter_3d}). As we can see from the figure, the performance of AGST goes up when we increase the value of $\lambda_1$ and reach the peak when $\lambda_1 = 1$. For $\lambda_2$, the best-performing value is $0.1$. The performance will decrease if the value is either too large or too small. We have aligned observations with other datasets. 

\begin{figure}[t]

    \graphicspath{{figures/}}
    \centering
    \subfigure 
    {
    \includegraphics[width=0.425\columnwidth]{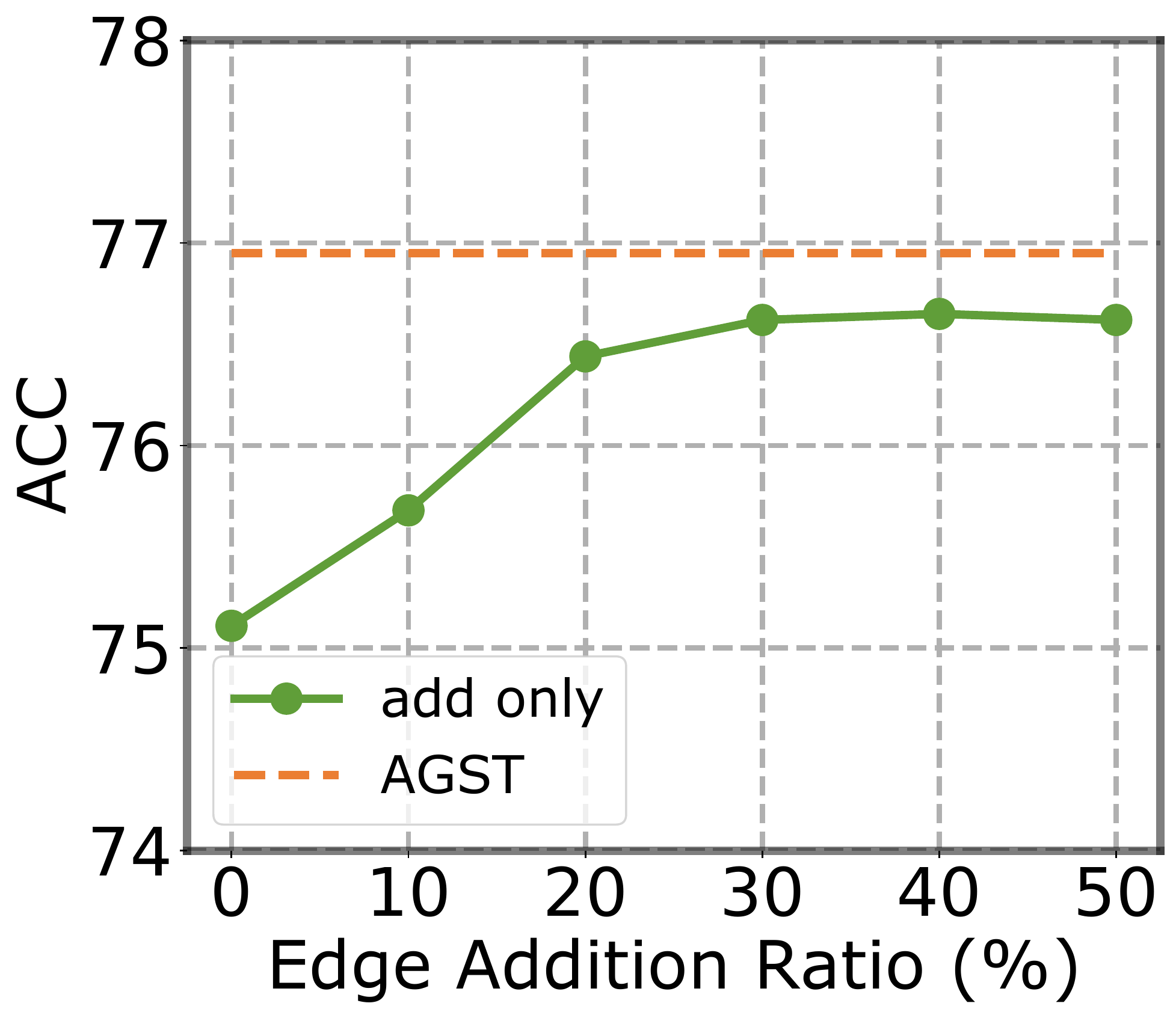}
    }
    \hspace{-0.1cm}
    \subfigure
    {
    \includegraphics[width=0.425\columnwidth]{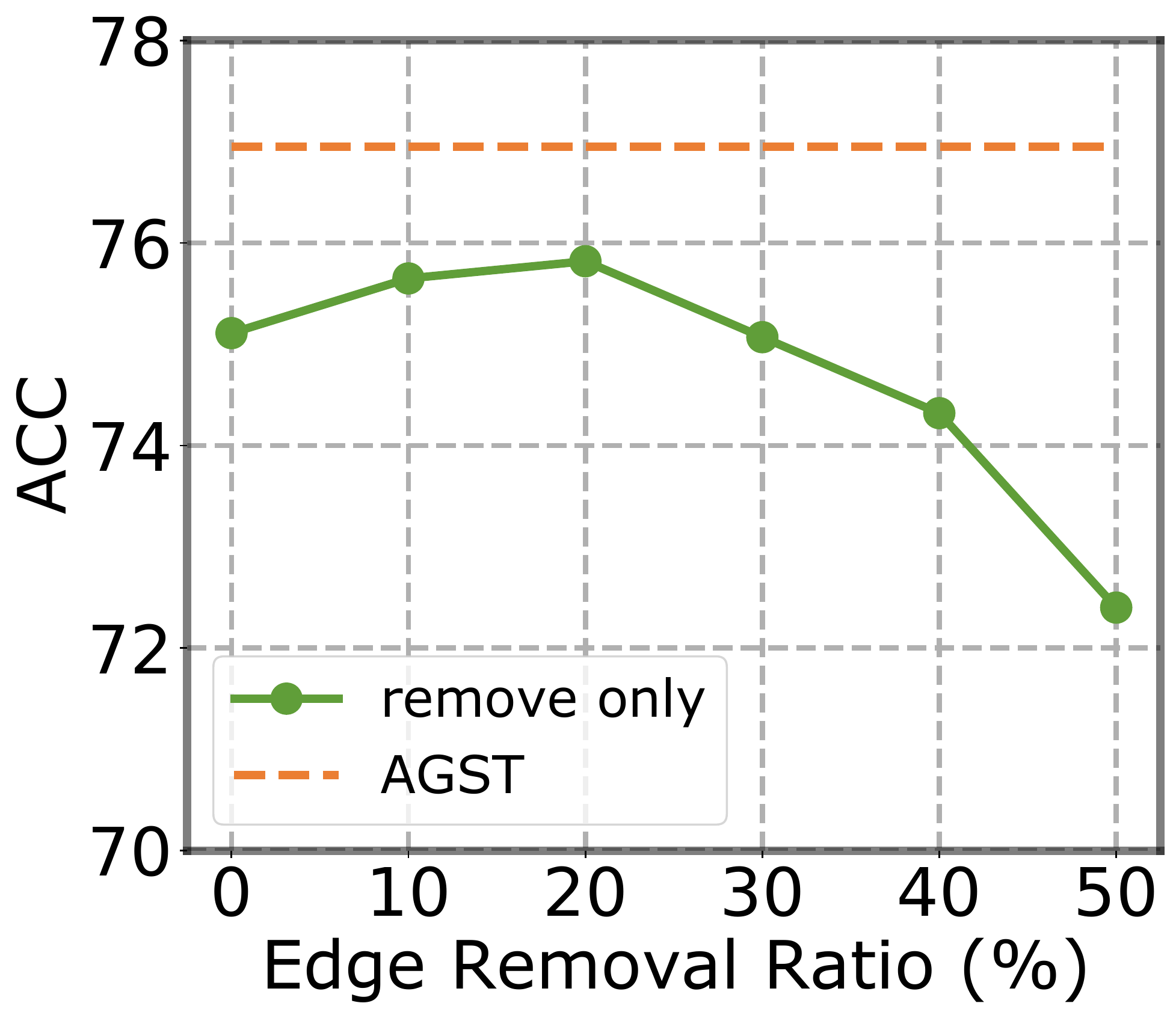}
    }
    \caption{Parameter analysis ($\beta_a, \beta_r$) results on Cora (5-shot).}%
    \label{fig:augment}
\end{figure}


\smallskip
\noindent\textbf{Topology Augmentation.} We further examine the impact of two hyper-parameters in our graph topology augmentation module, i.e. the edge addition ratio $\beta_a$ and the edge removal ratio $\beta_r$ for refining the graph structure. Figure \ref{fig:augment} shows the performance change (5-shot) on the Cora dataset by varying the value of each parameter, with a constant interval of 0.1. We also add the performance of AGST as reference. We observe that our graph topology augmentation function improves performance by either adding or removing edges within an appropriate range to mitigate the structure noise. For instance, the accuracy is first boosted by adding edges, then it reaches the peak until $\sim40\%$ and become stable throughout the range. Similarly, edge removal improves performance until $\sim20\%$, then the accuracy decreases quickly. One explanation is that a higher threshold may result in the accidental removal of possibly useful intra-class edges.





\section{Conclusion}

In this paper, we investigate the problem of semi-supervised node classification against data scarcity,  and introduce an augmented graph self-training framework (AGST) for the problem of semi-supervised node classification when labeled-node data is limited. Specifically, we introduce two unique graph data augmentation modules to capture structural and semantic information for graph self-training to improve the model performance using few labeled nodes. The empirical results over various benchmark datasets demonstrate the effectiveness of our proposed framework versus the baseline methods for classification with few labeled nodes. For future work, we are interested in exploring learnable graph augmentation module that can adapt to non-homophilous graphs and developing more powerful GST backbone model. 



\bibliographystyle{plain}
\bibliography{acmart}

\newpage
\appendix
\section{Appendix}

\subsection{Dataset and Implementation Details}
\label{detail}

\noindent\textbf{Datasets.} We adopt six benchmark datasets in our experiments. The detailed statistics of the datasets are summarized in Table \ref{table:datasets}.

\begin{table}[h!]
\caption{Summary statistics of the evaluation datasets.}
\centering
\scalebox{1.0}{
\begin{tabular}{@{}lccccccc@{}}
\toprule


\textbf{Dataset} & \# Nodes & \# Edges & \# Features & \# Classes \\ \midrule
Cora & 2,708 & 5,278 &  1,433 & 7 \\
CiteSeer & 3,327 & 4,552 & 3,703  & 6 \\
PubMed   & 19,717  &  44,324 &  500 & 3 \\
Coauthor-CS  & 18,333   & 81,894 & 6,805 & 15 \\
Coauthor-Physics  & 34,493   & 247,962 & 8,415 & 5 \\
Amazon-Photo  & 7,487   & 119,043 & 745 & 8 \\

\bottomrule
\end{tabular}}

\label{table:datasets}
\end{table}

\noindent\textbf{AGST.} We implement the proposed AGST in PyTorch with a 12 GB Titan Xp GPU. Specifically, we use two-layer MLP with 64 hidden units for the feature transformation module. The self-training iteration of AGST is set to $3$ for all the datasets. We optimize the model with Adam optimizer and grid search  for the edge addition/removal rate in $\{0.1, 0.2, 0.3, 0.4, 0.5, 0.6, 0.7, 0.8., 0.9, 1.0\}$. The optimal values are selected when the model achieve the best performance for validation set. The early stopping criterion uses a patience of $p = 100$ and an (unreachably high) maximum of $n = 10,000$ epochs. The patience is reset whenever the accuracy increases or the loss decreases on the validation set.


\smallskip
\noindent\textbf{Baselines.} In our experiments, we compare our approach with different methods including, LP, GCN, GAT, SGC, GLP, IGCN, CGPN, ST-GCNs, M3S and PTA. For the baseline methods, we adopt their public implementations and the details are as follows:
\begin{itemize}[leftmargin=*,noitemsep,topsep=1.5pt]
    \item \textbf{LP}~\cite{zhou2004learning}: For fair comparison, we use the same propagation step $T=10$ and the teleport probability as Meta-LP .
    \item \textbf{GCN}\footnote{ \url{https://github.com/tkipf/pygcn}} \cite{kipf2017semi}: We build the GCN model with two graph convolutional layers (64 dimensions) for learning node representations in the graph.
    \item \textbf{GAT}\footnote{\url{https://www.dgl.ai/}} \cite{velickovic2018graph}: The model consists of two graph attentional layers with 8 heads and the negative input slope of the LeakyReLU function is $0.2$ as suggested in the paper. 
    \item \textbf{SGC}\footnote{\url{https://github.com/Tiiiger/SGC}} \cite{wu2019simplifying}: After the feature pre-processing step, it learns the node representations with 2-layer feature propagation with 64 hidden units. 
    \item \textbf{GLP \& IGCN}\footnote{\url{https://github.com/liqimai/Efficient-SSL}} \cite{li2019label}:  It uses a two-layer structure (64 hidden units) in which the filter parameters $k$ and $\alpha$ is set to be 5 and 10 for 20-shot, and is set to be 10 and 20 for all the other tasks. The results with the best performing filter (i.e., RNM or AR) are reported.
   \item \textbf{CGPN}\footnote{\url{https://github.com/LEAP-WS/CGPN}}~\cite{wan2021contrastive}: We choose the CGPN-GAT instantiation due to its stable performance. Specifically, the graph poisson network is built with 2-layer neural network. For the graph neural network module, we use 2-layer GAT with 64 hidden units.

   
    \item \textbf{PTA}\footnote{\url{https://github.com/DongHande/PT_propagation_then_training}} \cite{dong2021equivalence}: For fair comparison, we use the same neural network model as AGST, which is a two-layer MLP with 64 hidden units. 
  \item \textbf{ST-GCNs}\footnote{\url{https://github.com/liqimai/gcn}} \cite{li2018deeper}: It represents four variants: \textbf{Co-train} , \textbf{Self-train}, \textbf{Union} and \textbf{Intersection}. We use ST-GCNs to report the results of the best performing framework. 
    \item \textbf{M3S}\footnote{\url{https://github.com/datake/M3S}} \cite{sun2020multi}:  We fix the number of clusters as 200 and select the best number of layers and stages as suggested by the authors. 

    
\end{itemize}

For all the baseline methods, we use Adam as optimizer and fine-tune the hyperparameters on each dataset. Specifically, we grid search for the learning rate in \{$1 \times 10^{-5}$, $5 \times 10^{-5}$, $1 \times 10^{-4}$, $5 \times 10^{-4}$, $1 \times 10^{-3}$, $5 \times 10^{-3}$, $1 \times 10^{-2}$, $5 \times 10^{-2}$, $1 \times 10^{-1}$, $5 \times 10^{-1}$ \} and dropout rate in $\{0.1, 0.2, 0.3, 0.4, 0.5, 0.6, 0.7\}$. Also, we use the same early stop strategy as for AGST. 

\smallskip
\noindent\textbf{Packages Used for Implementation.} For reproducibility, we also list the packages we use in the implementation with their corresponding versions: python==3.6.6, pytorch==1.4.0, cuda==10.1, torch-geometric==, numpy==1.19.2, dgl==0.6.1, and scikit-learn==0.24.0.

\subsection{Additional Results}
\label{appendix:results}
\begin{figure}[t]
    \graphicspath{{figures/}}
    \centering
    \subfigure[\textbf{Cora}]
    {
    \includegraphics[width=0.425\columnwidth]{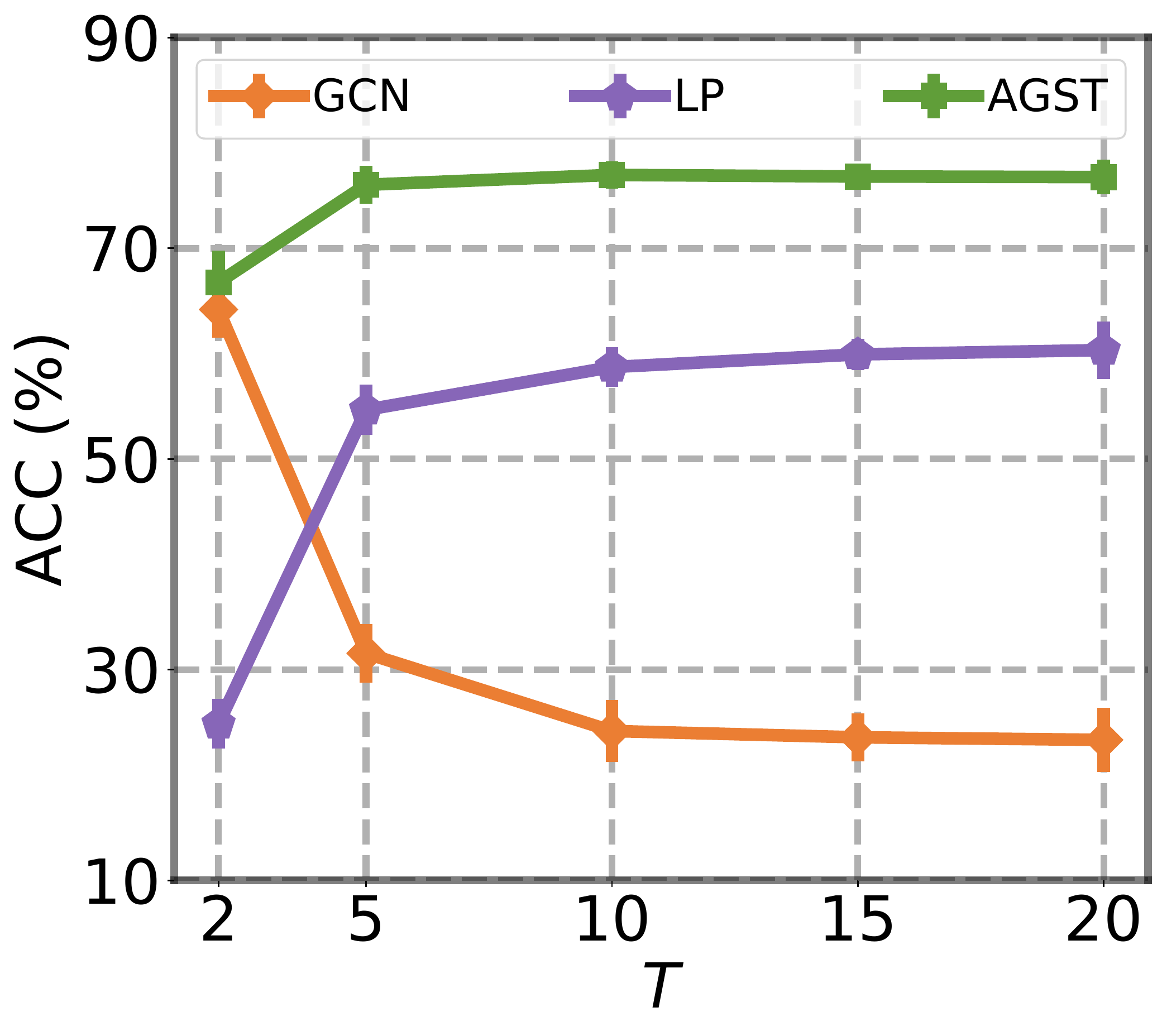}
    }
    \hspace{-0.1cm}
    \subfigure[\textbf{CiteSeer}]
    {
    \includegraphics[width=0.425\columnwidth]{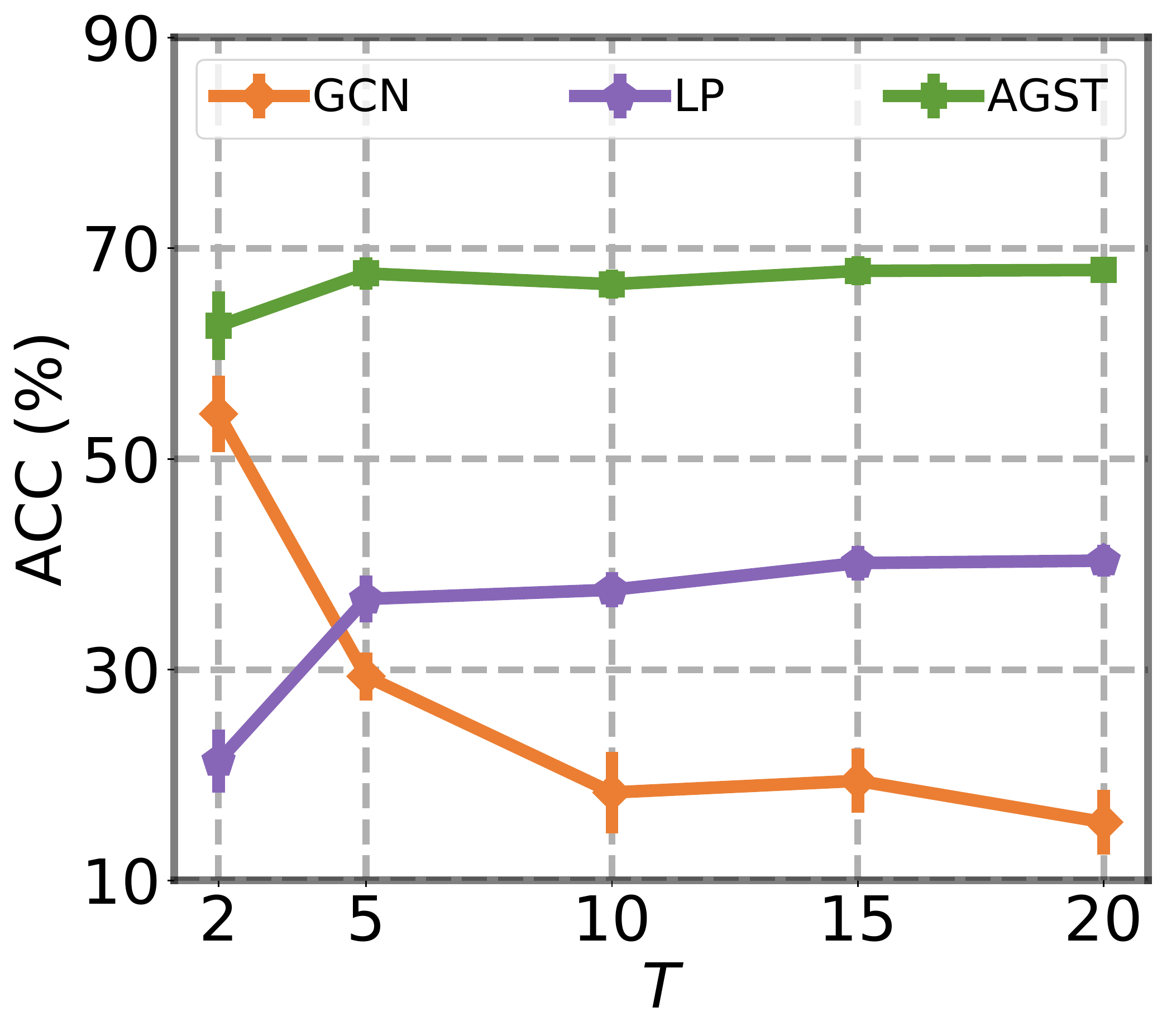}
    }
    \caption{Parameter analysis for propagation steps $\mathbf{T}$.}%

    \label{fig:parameter}
\end{figure}

\noindent\textbf{Propagation Steps.}
To demonstrate the effects of using different propagation steps, we compare our approach with two baselines (i.e., LP and GCN) under the $5$-shot setting with varying number of $T$. As shown in Figure \ref{fig:parameter}, we can clearly see that \textit{GCN} encounters performance degradation if we largely increase the number of propagation steps. Though LP is a naive baseline, its performance increases by using larger propagation steps. Our framework AGST adopts a decoupled backbone where label propagation serves as the teacher model, thus it is able to address the oversquashing issue and leverage large receptive fields. AGST achieve stable performance when the propagation step $T > 5$.

\end{document}